\documentclass[10pt,journal,compsoc]{IEEEtran}
\newif\ifpeerreview

\peerreviewfalse

\usepackage[switch]{lineno}

\usepackage{booktabs}       
\usepackage{amsfonts}       
\usepackage{nicefrac}       
\usepackage{microtype}      
\usepackage{xcolor}         

\usepackage{array}

\usepackage{bm}

\usepackage{graphicx}
\usepackage{pifont} 
\usepackage{subcaption} 

\usepackage{multirow}

\usepackage{algorithm}
\usepackage{setspace}
\usepackage{algorithmic}
\usepackage{amsmath}

\usepackage{xspace}
\makeatletter
\DeclareRobustCommand\onedot{\futurelet\@let@token\@onedot}
\def\@onedot{\ifx\@let@token.\else.\null\fi\xspace}

\def\eg{\emph{e.g}\onedot}

\def\etc{\emph{etc}\onedot}

\def\etal{\emph{et al}\onedot}
\makeatother

\AtEndPreamble{
    \usepackage[capitalize]{cleveref}
    \crefname{section}{Sec.}{Secs.}
    \Crefname{section}{Section}{Sections}
    \Crefname{table}{Table}{Tables}
    \crefname{table}{Tab.}{Tabs.}
}

\definecolor{darkpastelgreen}{rgb}{0.01, 0.75, 0.24}

\newcolumntype{H}{>{\setbox0=\hbox\bgroup}c<{\egroup}@{}}

\usepackage{fix-cm}

\definecolor{cvprblue}{rgb}{0.21,0.49,0.74}
\usepackage[pagebackref,breaklinks,colorlinks,allcolors=cvprblue]{hyperref}

\title{HiddenObject: Modality-Agnostic Fusion for Multimodal Hidden Object Detection}

\author{Harris Song$^{*}$,
        Tuan-Anh Vu$^{*}$,~\IEEEmembership{Member,~IEEE,}
        Sanjith Menon,
        Sriram Narasimhan,~\IEEEmembership{Member,~IEEE,}
        and M. Khalid Jawed,~\IEEEmembership{Member,~IEEE}
\IEEEcompsocitemizethanks{\IEEEcompsocthanksitem Harris Song is with the Department of Computer Science at the University of California, Los Angeles.\protect
\IEEEcompsocthanksitem Tuan-Anh Vu, Sanjith Menon, Sriram Narasimhan, M. Khalid Jawed are with the Department of Mechanical \& Aerospace Engineering at the University of California, Los Angeles.
\IEEEcompsocthanksitem $^{*}$ The authors contributed equally, with Tuan-Anh Vu serving as the project lead.
}
}

\begin{document}

\IEEEtitleabstractindextext{%
\begin{abstract}
Detecting hidden or partially concealed objects remains a fundamental challenge in multimodal environments, where factors like occlusion, camouflage, and lighting variations significantly hinder performance. Traditional RGB-based detection methods often \textit{fail under such adverse conditions}, motivating the need for more robust, modality-agnostic approaches. In this work, we present \textbf{HiddenObject}, a fusion framework that integrates RGB, thermal, and depth data using a Mamba-based fusion mechanism. Our method \textit{captures complementary signals across modalities}, enabling enhanced detection of obscured or camouflaged targets. Specifically, the proposed approach identifies modality-specific features and fuses them in a unified representation that generalizes well across challenging scenarios. We validate \textbf{HiddenObject} across multiple benchmark datasets, demonstrating \textit{state-of-the-art or competitive performance} compared to existing methods. These results highlight the efficacy of our fusion design and expose key limitations in current unimodal and naïve fusion strategies. More broadly, our findings suggest that Mamba-based fusion architectures can significantly advance the field of multimodal object detection, especially under visually degraded or complex conditions.
\end{abstract}

\begin{IEEEkeywords} 
hidden object, camouflaged object, occluded object, multimodal, object detection, feature fusion
\end{IEEEkeywords}
}


\maketitle

\IEEEtitleabstractindextext{%
\begin{abstract}
Detecting hidden or partially concealed objects remains a fundamental challenge in multimodal environments, where factors like occlusion, camouflage, and lighting variations significantly hinder performance. Traditional RGB-based detection methods often \textit{fail under such adverse conditions}, motivating the need for more robust, modality-agnostic approaches. In this work, we present \textbf{HiddenObject}, a fusion framework that integrates RGB, thermal, and depth data using a Mamba-based fusion mechanism. Our method \textit{captures complementary signals across modalities}, enabling enhanced detection of obscured or camouflaged targets. Specifically, the proposed approach identifies modality-specific features and fuses them in a unified representation that generalizes well across challenging scenarios. We validate \textbf{HiddenObject} across multiple benchmark datasets, demonstrating \textit{state-of-the-art or competitive performance} compared to existing methods. These results highlight the efficacy of our fusion design and expose key limitations in current unimodal and naïve fusion strategies. More broadly, our findings suggest that Mamba-based fusion architectures can significantly advance the field of multimodal object detection, especially under visually degraded or complex conditions.
\end{abstract}

\begin{IEEEkeywords} 
hidden object, camouflaged object, occluded object, multimodal, object detection, feature fusion
\end{IEEEkeywords}
}    

\IEEEraisesectionheading{
  \section{Introduction}\label{sec:introduction}
}
\label{sec:intro}

\IEEEPARstart{T}{he} rapid advancements in autonomous systems and robotic technologies have been driven by the need to enhance operational efficiency, mitigate labor shortages, and address complex real-world challenges across various domains, including industrial automation, agriculture, surveillance, search-and-rescue operations, and security applications. As these scenarios grow increasingly sophisticated, accurately detecting hidden or partially obscured objects becomes critical for effective decision-making and operational success. Traditional object detection methods primarily rely on single-modality imaging, notably RGB cameras, which are significantly limited by factors such as lighting variations, occlusions, camouflaged objects, and adverse environmental conditions~\cite{zou2023object,vu2023ovcis,vu2024transcues}.

Real-world objects often appear partially or fully concealed due to foliage, structural elements, environmental clutter, or intentional concealment, posing substantial detection challenges for conventional RGB imaging, which relies heavily on ideal lighting and unobstructed visibility. Conversely, multimodal imaging techniques—such as thermal~\cite{rai2017thermal} and depth~\cite{lopes2022survey} imaging—offer robust solutions by capturing complementary information even under challenging visual conditions. Thermal imaging, for instance, detects infrared radiation independent of visible light, demonstrating strong resilience against lighting variability and providing reliable performance in nighttime~\cite{Bijelic_2020_STF} or obscured environments~\cite{teng2023citrus}.

\begin{figure}[!t]
    \centering
    \includegraphics[width=0.99\linewidth]{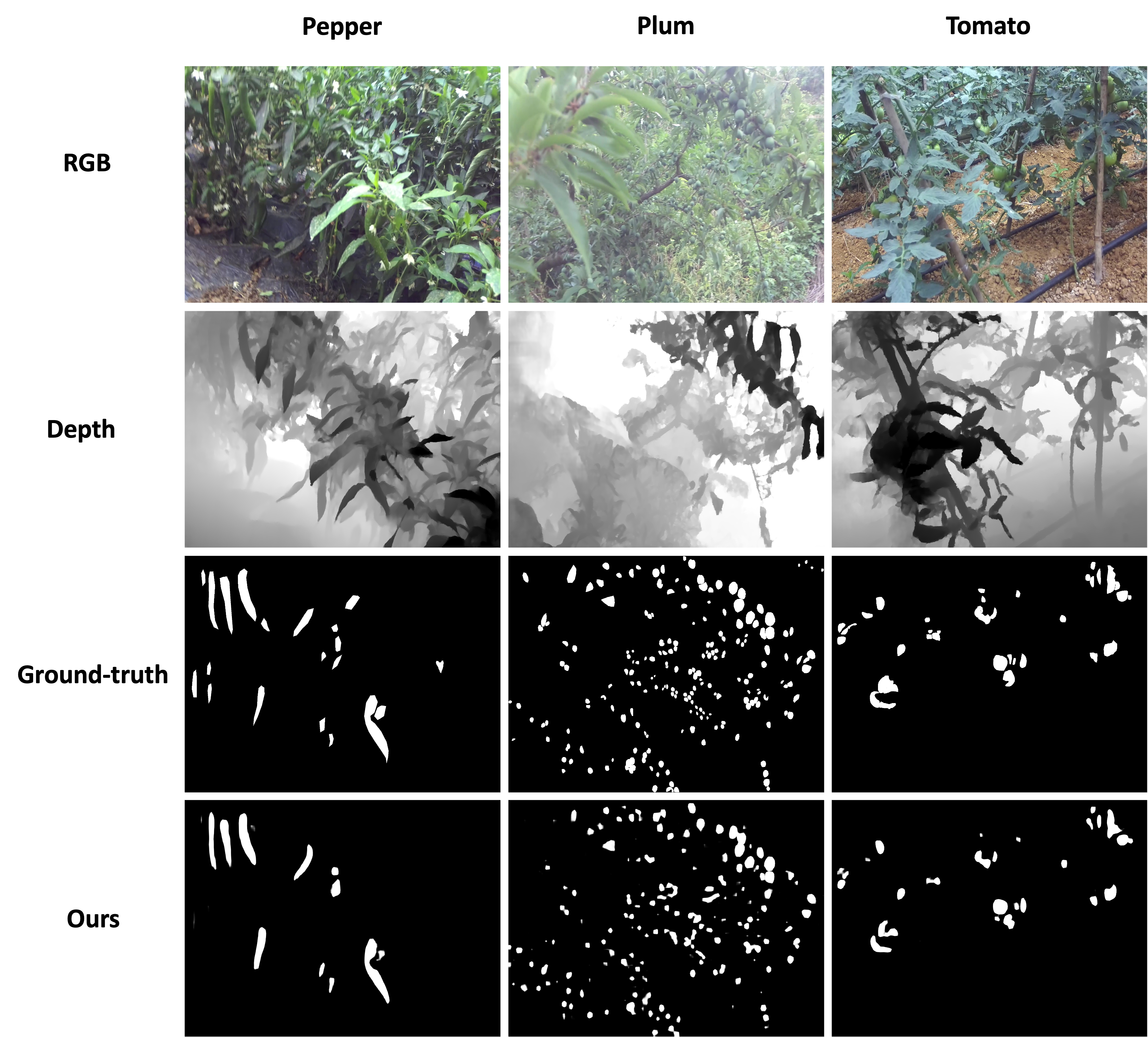}
    \caption{Our proposed method effectively addresses severe occlusion and densely distributed small objects, enhancing detection performance. The examples utilized are sourced from the ACOD-K12 dataset~\cite{Wang_2024_CVPR}.}
    \vspace{-1em}
    \label{fig:teaser}
\end{figure}

Integrating multiple imaging modalities thus presents significant advantages for handling complex detection tasks, including partial occlusions~\cite{Wang_2024_CVPR}, dense object clustering~\cite{CitDet}, and camouflage~\cite{prakash2021multi}. Objects exhibiting minimal color or texture contrast in RGB modalities often possess distinct signatures in thermal, depth, or near-infrared (NIR) modalities, thereby greatly enhancing detectability. Specifically, thermal imaging differentiates objects based on thermal contrast, while depth imaging captures structural three-dimensional information, further facilitating the detection of hidden or obscured objects~\cite{tang2024multi}.

\begin{figure*}[!t]
    \centering
    \includegraphics[width=0.99\linewidth]{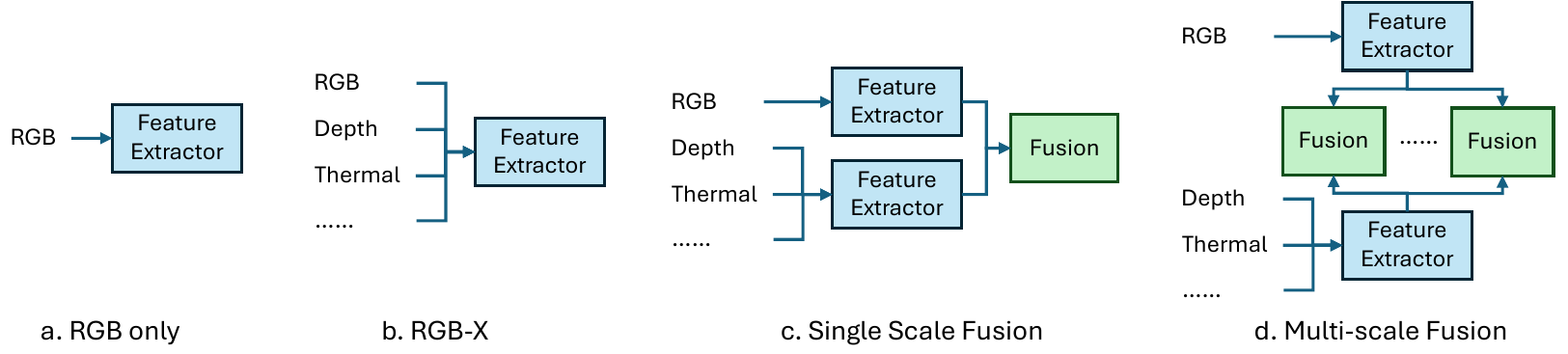}
    \caption{\textbf{Comparison of different multimodal fusion methods.} (a) Training a feature extractor using only RGB images. (b) Sharing a trainable feature extractor between RGB images and other modalities. (c) Single-scale feature fusion within the model. (d) Multiscale feature fusion (Our method).}
    \label{fig:evolution}
\end{figure*}

However, fusing multiple modalities into a cohesive detection framework introduces considerable technical challenges, such as modality misalignment, information redundancy, and computational efficiency. Existing approaches typically rely on static fusion strategies~\cite{ha2017mfet,sun2019rtfnet} or emphasize single modalities~\cite{shivakumar2020pst900}, which limits their ability to interpret critical information in complex or unfamiliar environments effectively (see~\Cref{fig:evolution}). To overcome these limitations, we introduce \textbf{HiddenObject}, a modality-agnostic detection framework utilizing a novel Mamba-based fusion mechanism coupled with a channel-aware decoder. Our approach efficiently extracts and seamlessly integrates critical information across diverse imaging modalities (e.g., RGB, thermal, depth), enabling robust and accurate detection of concealed objects in challenging scenarios.

As demonstrated in~\Cref{fig:teaser}, our framework maintains reliable detection performance even under severe occlusion and complex clutter conditions, a capability especially crucial in applications such as agriculture~\cite{Abdulsalam2023,firestereo,lee2024cart} and robotics~\cite{ha2017mfet}, where target objects frequently remain hidden or obscured. The key contributions of our work are summarized as follows:
\begin{itemize}
    \item We propose a Mamba-based fusion mechanism integrated with a channel-aware decoder to effectively extract and merge multimodal information from RGB, thermal, depth, and other modalities.
    \item Our detection pipeline robustly handles a wide range of hidden objects, including lightly obscured, partially occluded, heavily occluded, concealed, or camouflaged targets.
    \item Through extensive experiments across multiple datasets, we validate the effectiveness and efficiency of our proposed framework, providing initial insights into the applicability of Mamba for multimodal learning.
\end{itemize}

In the following sections, we provide detailed descriptions of our approach, including the fusion mechanism, dataset selection, experimental protocols, and comprehensive performance evaluations, further emphasizing the practical applicability and versatility of our proposed \textbf{HiddenObject} framework.

\section{Related Work}
\label{sec:related}

\begin{figure*}[!t]
\centering
\includegraphics[width=0.99\textwidth]{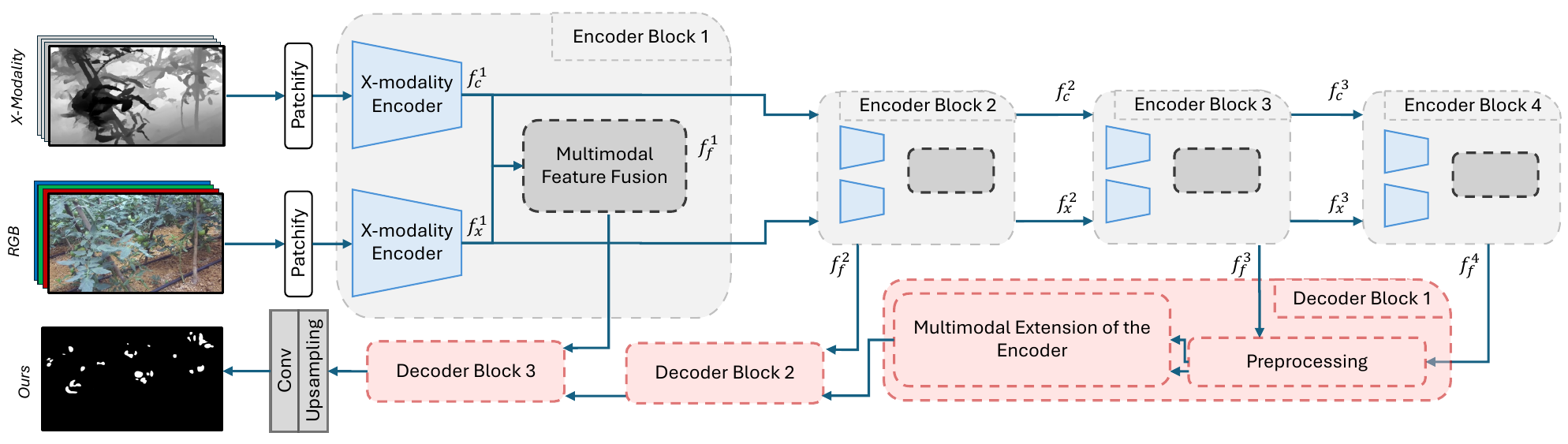}
\caption{Architecture overview of our proposed \textbf{HiddenObject} network, which integrates RGB and additional modality inputs for robust hidden object detection.}
\label{fig:arch}
\end{figure*}

\textbf{Transformer-based Object Detection.} Transformer architectures have significantly advanced object detection capabilities, starting notably with the DETR framework~\cite{detr}, which redefines object detection as a set prediction task, leveraging binary matching for one-to-one object associations during training. DETR eliminates the need for handcrafted anchor boxes and non-maximum suppression (NMS), yet it suffers from slow convergence. To mitigate this, several enhanced variants were developed: Deformable DETR~\cite{zhu2020deformable} accelerates training by introducing deformable attention mechanisms and reference anchor points; Conditional DETR~\cite{meng2021conditional} expedites convergence through conditional cross-attention separating content and positional information; Efficient DETR~\cite{yao2021efficient} integrates dense detection with sparse predictions to improve efficiency; DAB-DETR~\cite{liu2022dab} refines bounding box predictions through iterative 4D reference points; DN-DETR~\cite{li2022dn} incorporates query denoising for improved training efficiency; and DINO~\cite{zhang2022dino} consolidates these improvements into a unified detection pipeline. Real-time implementations, such as RT-DETR~\cite{rtdetrv1,rtdetrv2}, further enhance practicality using Efficient Hybrid Encoders and IoU-aware query selection strategies.

\vspace{0.8em}
\noindent \textbf{State Space Models (SSMs) in Vision.} Recent developments in SSMs, particularly the Mamba architecture~\cite{gu2024mamba}, provide linear-time complexity beneficial for sequence modeling tasks, overcoming transformer limitations on long sequence handling. Vision-specific adaptations, such as Vision Mamba (Vim)~\cite{zhu2024vision}, significantly improve computational efficiency and reduce memory usage compared to traditional Vision Transformers (\eg DeiT~\cite{touvron2021training}). Integrations into object detection frameworks, such as Mamba YOLO~\cite{wang2025mamba} and Fusion-Mamba~\cite{dong2024fusion}, have demonstrated significant accuracy improvements in multimodal scenarios, highlighting their potential in concealed object detection. Coupled Mamba~\cite{coupled2024mamba} and Sigma~\cite{wan2025sigma} further enhance multimodal fusion performance, providing improved cross-modal consistency and inference speed, indicating its suitability for complex detection tasks involving hidden or obscured objects.

\vspace{0.8em}
\noindent \textbf{Multimodal Detection.} Earlier infrared-visible detection studies typically adapted existing single-modality frameworks, such as the Faster R-CNN~\cite{ren2015faster} and YOLO detectors~\cite{redmon2016you,redmon2017yolo9000,wang2023yolov7}. König \etal \cite{konig2017fully} introduced a fully convolutional fusion Region Proposal Network (RPN), demonstrating the advantage of mid-level feature fusion. Subsequent efforts developed CNN-based attention mechanisms to enhance cross-modal synergy~\cite{qingyun2022cross,roszyk2022adopting,cao2023multimodal}. More recent transformer-based fusion strategies~\cite{qingyun2021cross,fu2023lraf,zhu2023multi,shen2024icafusion} exploit broader complementary interactions between infrared and visible data. Other approaches incorporate global illumination data as weights for fusion or output refinement to mitigate noise effects~\cite{zhou2020improving,yang2022baanet}. Recognizing regional complementarity variations, segmentation-based bounding-box-level fusion~\cite{zhang2020multispectral,zhang2021guided} and ROI prediction methods~\cite{zhang2021weakly,kim2021uncertainty} were explored. Methods utilizing confidence or uncertainty metrics also refine multi-branch predictions post-fusion~\cite{li2022confidence,li2023stabilizing}. Addressing modality misalignment, Zhang \etal \cite{zhang2021weakly} proposed AR-CNN for feature alignment across modalities, and Kim \etal \cite{kim2021mlpd} employed multi-label learning to enhance detection robustness. More recently, multimodal detection methods such as DAMSDet~\cite{guo2024damsdet} dynamically adapt fusion strategies, and Zhang \etal \cite{zhang2024e2e} presented a novel end-to-end multimodal fusion detection pipeline, highlighting continuous advancement in this field.
\section{Proposed Method}
\label{sec:method}

This section introduces the proposed Multimodal Mamba-based detection framework, as shown in~\Cref{fig:arch}, which consists of three main components: a dual-stream feature extraction backbone network based on a two-dimensional selective state-space model, a cross-modal feature fusion module, and a channel-wise Mamba decoder. Unlike prior methods that use modality-specific pathways or require retraining for different modality combinations, our model can flexibly handle varying modality inputs without requiring architectural changes. This, combined with the SS2D design, enables efficient fusion and enhances generalization in multimodal scenarios. We will first introduce the fundamental concepts of the visual selective state-space model in~\Cref{subsec:pre}. Then, in~\Cref{subsec:enc}, we will describe the dual-stream Mamba-based encoder. Next, in~\Cref{subsec:fusion} will cover the proposed multimodal fusion module. Finally, the channel-wise Mamba decoder is discussed in~\Cref{subsec:dec}.

\begin{figure*}[!t]
\centering
\includegraphics[width=0.99\linewidth]{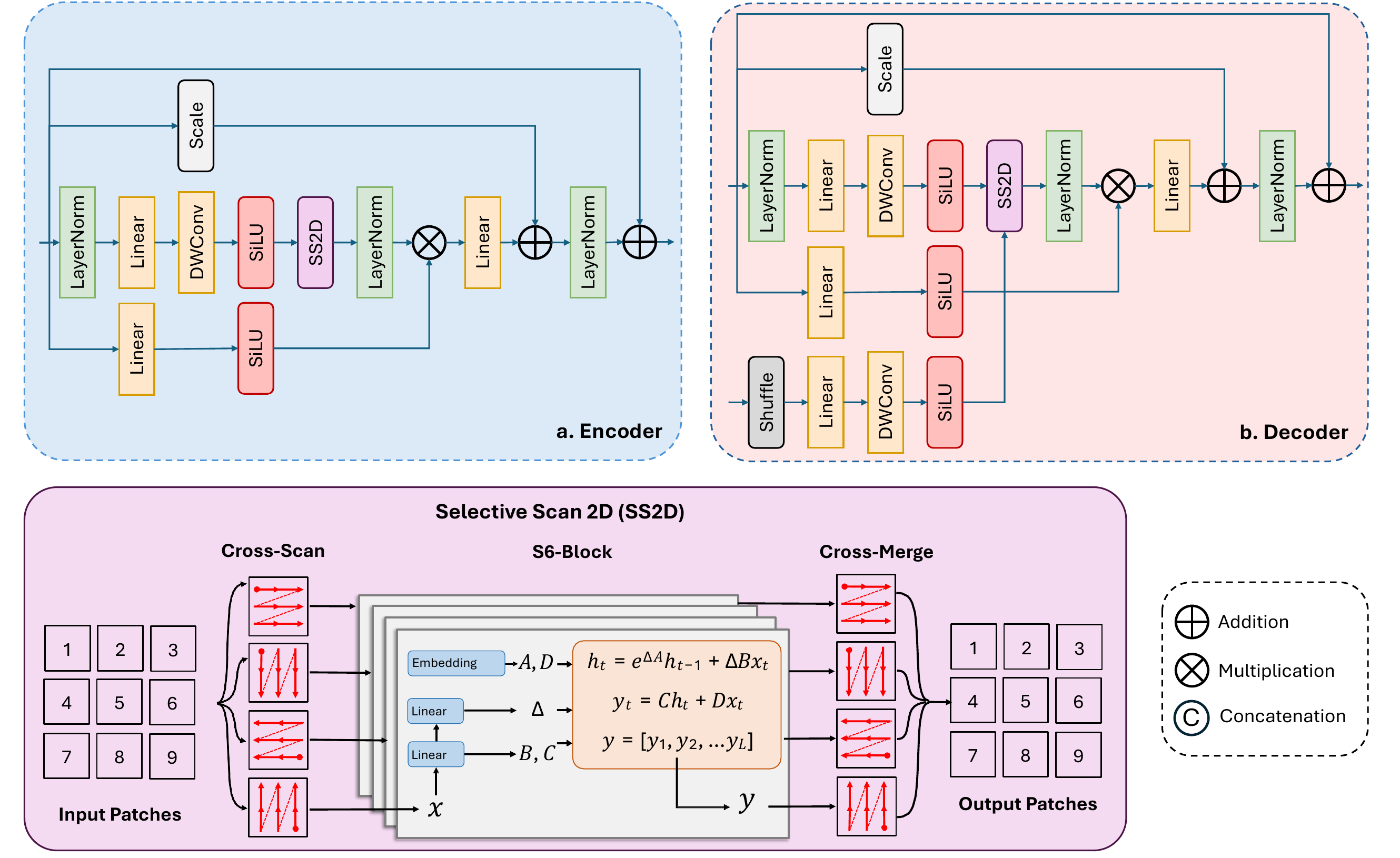}
\caption{\textbf{The architecture of Vision Mamba Encoder and Decoder.}
For the SS2D module in the decoder, the matrices $A, B,$ and $\Delta$ are computed from the X-modality, the input of the selective scan is the X-modality embedding, and the matrix $C$ is calculated using the RGB feature.
}
\label{fig:encoder_decoder}
\end{figure*}

\subsection{Preliminaries}
\label{subsec:pre}

State Space Models (SSM)~\cite{gu2022efficiently,gu2021combining,smith2023simplified} constitute a type of sequence-to-sequence model characterized by time-invariant dynamics, referred to as linear time-invariance. Due to their linear computational complexity, SSMs effectively capture dynamic patterns by implicitly mapping inputs to latent states, but scale linearly with respect to the sequence length, formulated as:
\begin{equation}
  \label{eq:ssm}
  y(t) = Ch(t) + Dx(t), \quad
  \dot{h}(t) = Ah(t) + Bx(t).
\end{equation}

Here, $x(t) \in \mathbb{R}$ denotes the input, $h(t) \in \mathbb{R}^{N}$ the hidden state, and $y(t) \in \mathbb{R}$ the output. The term $\dot{h}(t)$ represents the temporal derivative of $h(t)$, and $N$ denotes the dimensionality of the state. Matrices ${A} \in \mathbb{R}^{N \times N}$, ${B} \in \mathbb{R}^{N \times 1}$, ${C} \in \mathbb{R}^{1 \times N}$, and scalar ${D} \in \mathbb{R}$ define the system dynamics. For discrete sequences such as images and text, SSMs utilize Zero-Order Hold (ZOH) discretization~\cite{gu2022efficiently} to map an input sequence $\{x_1, x_2, \dots, x_K\}$ to an output sequence $\{y_1, y_2, \dots, y_K\}$. Given a predefined timescale parameter $\mathrm{\Delta} \in \mathbb{R}^{D}$, the continuous parameters $A$ and $B$ are discretized as follows:
\begingroup
\begin{gather}
\overline{A} = \exp(\mathrm{\Delta} A), \quad \overline{B} = (\mathrm{\Delta} A)^{-1}(\exp(A)-I)\mathrm{\Delta} B, \quad \overline{C} = C, \\
y_k = \overline{C}h_k + \overline{D}x_k, \quad
h_k = \overline{A}h_{k-1} + \overline{B}x_k.
\label{eq:ss}
\end{gather}
\endgroup

All matrices retain their dimensions throughout the iterative computations. Typically, the residual connection term $\overline{D}$ is omitted, simplifying the equation to:
\begin{equation}
    y_k = \overline{C}h_k.
    \label{eq:y=ch}
\end{equation}
Additionally, following the approach of Mamba~\cite{gu2024mamba}, the matrix $\overline{B}$ is approximated using a first-order Taylor expansion:
\begin{equation}
    \overline{B} = (\exp(A)-I)A^{-1}B \approx (\mathrm{\Delta} A)(\mathrm{\Delta} A)^{-1}\mathrm{\Delta} B = \mathrm{\Delta} B.
\end{equation}

\vspace{0.8em}
\noindent \textbf{Selective Scan Mechanism.}
Although effective in handling discrete sequences, standard SSMs contain parameters invariant to input changes. To overcome this limitation, the Selective State Space Model (S6, \textit{a.k.a} Mamba)~\cite{gu2024mamba} was proposed to introduce input-dependent parameters. In Mamba, the matrices $B \in \mathbb{R}^{L \times N}$, $C \in \mathbb{R}^{L \times N}$, and $\mathrm{\Delta} \in \mathbb{R}^{L \times D}$ are dynamically derived from the input $x \in \mathbb{R}^{L \times D}$, thus making the model responsive to the context of the input. This selective scanning mechanism enables Mamba to capture the intricate dependencies within long sequences efficiently.

\subsection{Mamba-based Dual-Stream Encoder}
\label{subsec:enc}

Inspired by recent methods on dual stream encoder~\cite{zhang2023cmx,gemenifusion,wan2025sigma}, our Mamba-based Dual-Stream Encoder is used to encode RGB and X-Modality (Thermal, Depth, NIR, \etc). We apply SS2D modeling with a carefully designed sequence of depth-wise convolutions and linear layers that enhance modality-agnostic fusion while maintaining efficiency. The encoder consists of two branches with shared weights, each processing an RGB image and an X-modality image, denoted as $I_{\mathsf{rgb}}, I_{\mathsf{x}} \in \mathbb{R}^{H \times W \times 3}$, where $H$ and $W$ represent the height and width of the input images, respectively. The input images are initially divided into patches and projected into 1D embeddings. These features are then processed through four successive blocks (Encoder Blocks), each comprising X-modality Vision Mamba~\cite{liu2024vmamba} encoders and a Multimodal Feature Fusion (MMFF) module, to produce hierarchical multiscale features.

\vspace{0.8em}
\noindent \textbf{Vision Mamba Encoder.} 
Following the VMamba framework, the encoder incorporates Selective Scan 2D (SS2D) modules. As illustrated in~\Cref{fig:encoder_decoder}, the input features sequentially undergo Layer Normalization (LN), a linear projection, depth-wise convolution, SiLU layer, and 2D-Selective-Scan (SS2D) module. We follow~\cite{liu2024vmamba} and use four different scan directions to maintain spatial information. Then, for each scan, we calculate the state and output following~\Cref{eq:ssm} and merge them together by reordering and summing them.

\vspace{0.8em}
\noindent \textbf{SS2D Module.} 
The SS2D module operates in three stages: cross-scan, selective scanning with S6 blocks, and cross-merge. Initially, input features are reorganized into four directional sequences (top-left to bottom-right, bottom-right to top-left, top-right to bottom-left, and bottom-left to top-right). Distinct selective scan modules independently process each sequence to capture comprehensive multi-directional long-range dependencies~\cite{gu2024mamba}. Finally, the cross-merge step aggregates and merges the sequences via summation, restoring the 2D feature structure.

\subsection{Our Multimodal Feature Fusion}
\label{subsec:fusion}

Inspired by CMX~\cite{zhang2023cmx} and Sigma~\cite{wan2025sigma}, we designed our feature fusion module to enable interactive, bidirectional cross-modal feature rectification and sequence-to-sequence cross-attention. This approach facilitates comprehensive cross-modal interactions~\cite{wan2025sigma}, moving beyond simple input merging through modality-specific operations~\cite{cao2021shapeconv,chen2021spatial} or unidirectional fusion using channel attention~\cite{hu2019acnet,chen2020bi}. \Cref{fig:evolution} shows the comparison of different multimodal fusion methods. Our Fusion Module is distinct in: \ding{182} its design is tailored for flexibility in missing-modality scenarios, \ding{183} the specific arrangement and interaction of the SS2D blocks for each modality, and \ding{184} the scale-aware feature alignment before fusion, which is not found in Sigma~\cite{wan2025sigma}.

\begin{figure*}[!t]
\centering
\includegraphics[width=0.99\linewidth]{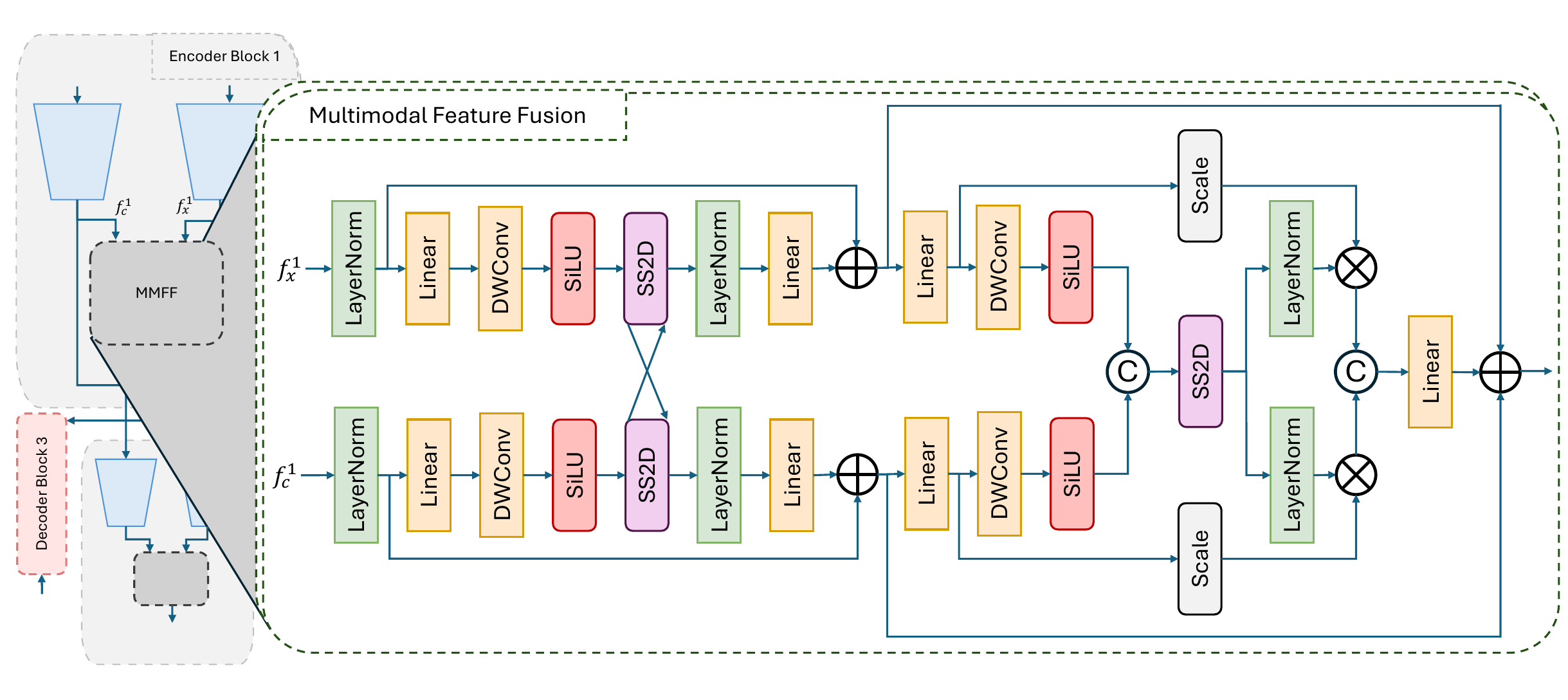}
\caption{Multimodal Feature Fusion (MMFF) Module.}
\label{fig:fusion}
\end{figure*}

As shown in~\Cref{fig:fusion}, two input features are initially processed through linear layers and depth-wise convolutions before being cross-passed to the selective scan. Following the selection mechanism of Mamba described in~\Cref{subsec:pre}, the system matrices $A, B,$ and $\Delta$ are generated to enhance the model's context-aware capabilities. Linear projection layers are employed to create these matrices. Matrix $C$ decodes information from the hidden state to produce the output, facilitating information exchange among the various selective scan modules. We scan the concatenated sequence in reverse to effectively capture information from both modalities and generate an additional sequence. Each sequence is then processed to identify long-range dependencies. The output from the inverted sequence is flipped back and added to the processed sequence. Next, the combined sequence is separated to recover the two outputs. After obtaining the scanned features, they are multiplied by two scaling parameters and concatenated in channel dimension. Finally, a linear projection layer is used to reduce feature shape.

\begin{table*}[!t]
\caption{Comparison between different datasets in our experiments.}
\label{tab:datasets}
\centering
\small
\setlength{\tabcolsep}{4pt}
\renewcommand{\arraystretch}{1.4}
\resizebox{0.99\linewidth}{!}{%
\begin{tabular}{@{}c|c|ccc|ccccc@{}}
\toprule
\multirow{2}{*}{\textbf{Dataset}} & \multirow{2}{*}{\textbf{\# Images}} & \multicolumn{3}{c|}{\textbf{Modality}} & \multicolumn{5}{c}{\textbf{Environments}} \\ \cmidrule(l){3-10} 
                                  &                                     & \textbf{RGB} & \textbf{Depth} & \textbf{Thermal} & \textbf{Concealed} & \textbf{Occluded} & \textbf{Nighttime} & \textbf{Interior} & \textbf{Precipitation} \\ \midrule \midrule
ACOD-12K~\cite{Wang_2024_CVPR}                          & 12,148                              & \checkmark   & \checkmark     &                   & \checkmark         & \checkmark        & \checkmark         &              &  \checkmark        \\
MFNet~\cite{ha2017mfet}                              & 1,569                                & \checkmark   &                & \checkmark        & \checkmark         & \checkmark        & \checkmark         &                         &                     \\
PST900~\cite{shivakumar2020pst900}                            & 894                                  & \checkmark   &                & \checkmark        & \checkmark         & \checkmark        &                     &                         &                     \\
NYU Depth V2~\cite{Silberman2012}                      & 1,449                                & \checkmark   & \checkmark     &                   &                     &                    &                     &  \checkmark                        &          \\
SUN RGB-D~\cite{SunRGBD}                         & 10,335                               & \checkmark   & \checkmark     &                   &                     &                    &                     &   \checkmark                       &          \\
\bottomrule
\end{tabular}
}
\end{table*}

\subsection{Mamba-based Decoder}
\label{subsec:dec}

As illustrated in~\Cref{fig:encoder_decoder}, the Mamba-based Decoder processes spatial-temporal features from MMFF modules at multiple hierarchical levels. Our decoder integrates modality-specific and shared information using a distinct update path and attention formulation compared to Sigma~\cite{wan2025sigma}. Structurally similar to the encoder above, the decoder operates through four stages with progressively varying channels, accommodating dual inputs. Lower-level features ($f^n_f$) undergo random shuffling and upsampling for enhanced information extraction. For the SS2D module in the decoder, the matrices $A, B,$ and $\Delta$ are computed from the lower-level features, the input of the selective scan is the lower-level embedding, and the matrix $C$ is calculated using the higher-level feature ($f^{n-1}_f$).

\section{Experiments}
\label{sec:exp}

To evaluate the effectiveness of our multimodality Mamba-based vision model, we benchmarked it against a diverse set of state-of-the-art thermal and depth-based object detection architectures. Standard evaluation metrics, including pixel-level and structural accuracy measures, were employed to ensure a fair comparison. Our analysis highlights the quantitative strengths and potential weaknesses of our model across multiple challenging datasets and modalities.

\subsection{Experimental Settings}
\label{sec:implementation}

\noindent\textbf{Baselines.} We compare our Mamba-based model against a diverse set of state-of-the-art RGB-Thermal and RGB-Depth fusion architectures to evaluate cross-modal robustness and generalization. These baselines address key challenges, including modality alignment, efficient fusion, and resilience under adverse conditions such as occlusion and low visibility. Our selection ensures broad coverage of both architectural paradigms and task settings, enabling a rigorous evaluation of the proposed model across real-world scenarios. 

\begin{itemize}
    \item \textbf{RGB-Thermal Fusion Models:} MFNet~\cite{ha2017mfet}, RTFNet~\cite{sun2019rtfnet}, and PSTNet~\cite{shivakumar2020pst900} serve as foundational RGB-T fusion networks with early and late fusion strategies. GMNet~\cite{zhou2021gmnet}, CACFNet~\cite{yi2022ccaffmnet}, and CAINet~\cite{lv2023cainet} introduce global and cross-attention mechanisms for adaptive fusion. LASNet~\cite{li2022rgb} focuses on lightweight and efficient fusion, making it suitable for real-time or constrained applications.
    \item \textbf{RGB-Depth Fusion Models:} Transformer-based methods like TokenFusion~\cite{wang2022multimodal} and MultiMAE~\cite{bachmann2022multimae} model long-range interactions. CNN-based architectures, including CEN~\cite{wang2020deep}, PDCNet~\cite{yang2023pixel}, and CMNeXt~\cite{zhang2023delivering}, emphasize depth-aware feature extraction and spatial context modeling. CAINet, originally focused on thermal applications, has also been evaluated for its cross-modal capabilities in RGB-D tasks.
    \item \textbf{Agricultural RGB-Depth Fusion Models:} We benchmark against recent RGB-D saliency detection models: DaCOD~\cite{wang2023depth}, PopNet~\cite{wu2023source}, HitNet~\cite{hu2023high}, FSPNet~\cite{huang2023feature}, HIDANet~\cite{wu2023hidanet}, XMSNet~\cite{wu2023object}, RISNet~\cite{Wang_2024_CVPR}, and FusionMamba~\cite{xie2024fusionmamba}, which tackle concealment and fine object segmentation through guided attention, hierarchical refinement, and transformer-based fusion.
\end{itemize}

\noindent\textbf{Implementation Detail.} 
We conduct three evaluations on the ACOD-12K dataset: (1) RGB–thermal (RGB-T), (2) RGB–depth (RGB-D), and (3) RGB-D on the concealed-object subset. Although the RGB-T and RGB-D tracks are often benchmarked with semantic-segmentation metrics, Camouflaged Object Detection (COD) typically uses a distinct evaluation protocol. Because ACOD-12K contains a higher proportion of heavily concealed and occluded instances, we adopt the standard COD metrics for reporting results. Following CMX~\cite{zhang2023cmx} and Sigma~\cite{wan2025sigma}, we use the AdamW optimizer with an initial learning rate of $6\times10^{-5}$ and weight decay of $0.01$. Models are trained for 500 epochs with a batch size of 8, using the ImageNet-1K–pretrained VMamba as the image encoder. All training and inference use an input resolution of $640\times480$. All methods were trained and evaluated on a cluster equipped with NVIDIA RTX 3090 GPUs. While the cluster has eight RTX 3090 GPUs, all results reported in this paper were obtained using four GPUs.

\vspace{0.8em}
\noindent\textbf{Dataset Overview.} As summarized in ~\ref{tab:datasets}, our experiments span five datasets: a concealed fruit detection dataset (ACOD-12K~\cite{Wang_2024_CVPR}), two publicly available RGB-Thermal (RGB-T) semantic segmentation datasets (MFNet~\cite{ha2017mfet}, PST900~\cite{shivakumar2020pst900}), two RGB-Depth (RGB-D) datasets (NYU Depth V2~\cite{Silberman2012} and SUN RGB-D~\cite{SunRGBD}). These datasets offer diverse sensing modalities (RGB, Depth, Thermal) and challenging real-world environments, including occlusion, concealment, low-light/nighttime conditions, precipitation, and complex indoor scenes. This diversity ensures rigorous testing of both the multimodal fusion capability and the model’s generalizability.

\vspace{0.8em}
\noindent\textbf{Evaluation Metrics.} We adopt standard evaluation metrics widely used in salient object detection, including the structure-measure \(S_\alpha\) ~\cite{fan2017structure}, the enhanced-alignment measure \(E_\phi\) introduced in~\cite{Fan2018EnhancedAlignment}, and the manifold ranking-based saliency formulation \textbf{$F_{\beta}^w$} ~\cite{Cheng2013ManifoldRanking}. The Mean Accuracy (mAcc) and Mean Intersection over Union (mIoU) methods were calculated for MFNet and PST900, with a focus on semantic segmentation and pixel-level correctness.

\begin{table}[!t]
\caption{Quantitative Comparisons on the MFNet and PST900 Datasets (RGB-Thermal).}
\label{tab:thermal}
\centering
\small
\setlength{\tabcolsep}{3pt}
\renewcommand{\arraystretch}{1.4}
\resizebox{\linewidth}{!}{%
\begin{tabular}{c|c|cc|cc}
\toprule
\multirow{2}{*}{\textbf{Method}} & \multirow{2}{*}{\textbf{Backbone}} & \multicolumn{2}{c|}{\textbf{MFNet}} & \multicolumn{2}{c}{\textbf{PST900}} \\ \cmidrule(l){3-6} 
                                 &                                    & \textbf{mAcc}$\uparrow$    & \textbf{mIoU}$\uparrow$    & \textbf{mAcc}$\uparrow$    & \textbf{mIoU}$\uparrow$    \\ 
                                 \midrule \midrule
MFNet~\cite{ha2017mfet}                & -                                  & 45.1             & 39.7             & -                & 57.0             \\
RTFNet~\cite{sun2019rtfnet}            & ResNet-152                         & 63.1             & 53.2             & -                & 57.6             \\
PSTNet~\cite{shivakumar2020pst900}     & ResNet-18                          & -                & 48.4             & -                & 68.4             \\
GMNet~\cite{zhou2021gmnet}             & ResNet-50                          & 74.1             & 57.3             & 89.6             & 84.1             \\
CACFNet~\cite{yi2022ccaffmnet}         & ConvNeXt-B                         & -                & 57.8             & -                & 86.6             \\
LASNet~\cite{li2022rgb}                & ResNet-152                         & 75.4             & 54.9             & 91.6             & 84.4             \\
CAINet~\cite{lv2023cainet}             & MobileNet-V2                       & 73.2             & 58.6             & 94.3             & 84.7             \\
CMX~\cite{zhang2023cmx}                & MiT-B4                             & -                & 59.7             & -                & -             \\
Sigma~\cite{wan2025sigma}              & VMamba-S                           & 73.3             & 61.1             & 92.6             & 87.8              \\ 
Ours                                   & VMamba-S                           & 73.8             & 61.5             & 93.3             & 88.5              \\ 
\bottomrule
\end{tabular}
}
\end{table}

\subsection{Benchmarks and Discussions}
\label{sec:benchmark}

We conduct comprehensive evaluations across multiple datasets and modalities to benchmark the performance of our proposed multimodal Mamba-based model. From the depth-based datasets, our Mamba-based approach achieved the highest mIoU for the SUN RGB-D modality algorithm, while the NYU Depth V2 dataset was only surpassed by the MiT-B4 algorithm. On the MFNet and PST900 RGB-T datasets, we have the highest mIoU in both datasets and have high median accuracy, although it is surpassed by MobileNet-V2 for the PST900 dataset and LASNet for the MFNet dataset. Our proposed model, based on the lightweight and efficient VMamba-S backbone, outperforms or closely matches top-performing models across all metrics, particularly in $F_{\beta}^w$ and $E_{\phi}$, demonstrating strong capability in capturing structural and spatial saliency from multimodal cues.

\vspace{0.8em}
\noindent\textbf{RGB-Thermal Experiments.} In the RGB-Thermal domain, our model also demonstrates state-of-the-art performance. On the MFNet~\cite{ha2017mfet}, PST900~\cite{shivakumar2020pst900} datasets (\Cref{tab:thermal}), our model achieves the \textbf{highest mIoU} of \textbf{61.5} on MFNet and \textbf{88.5} on PST900, while maintaining strong mAcc scores of \textbf{73.8} and \textbf{93.3}, respectively. Although CAINet slightly outperforms us in mAcc on PST900, our model remains superior in overall segmentation quality (mIoU).

\begin{figure*}[!t]
\centering
\includegraphics[width=\linewidth]{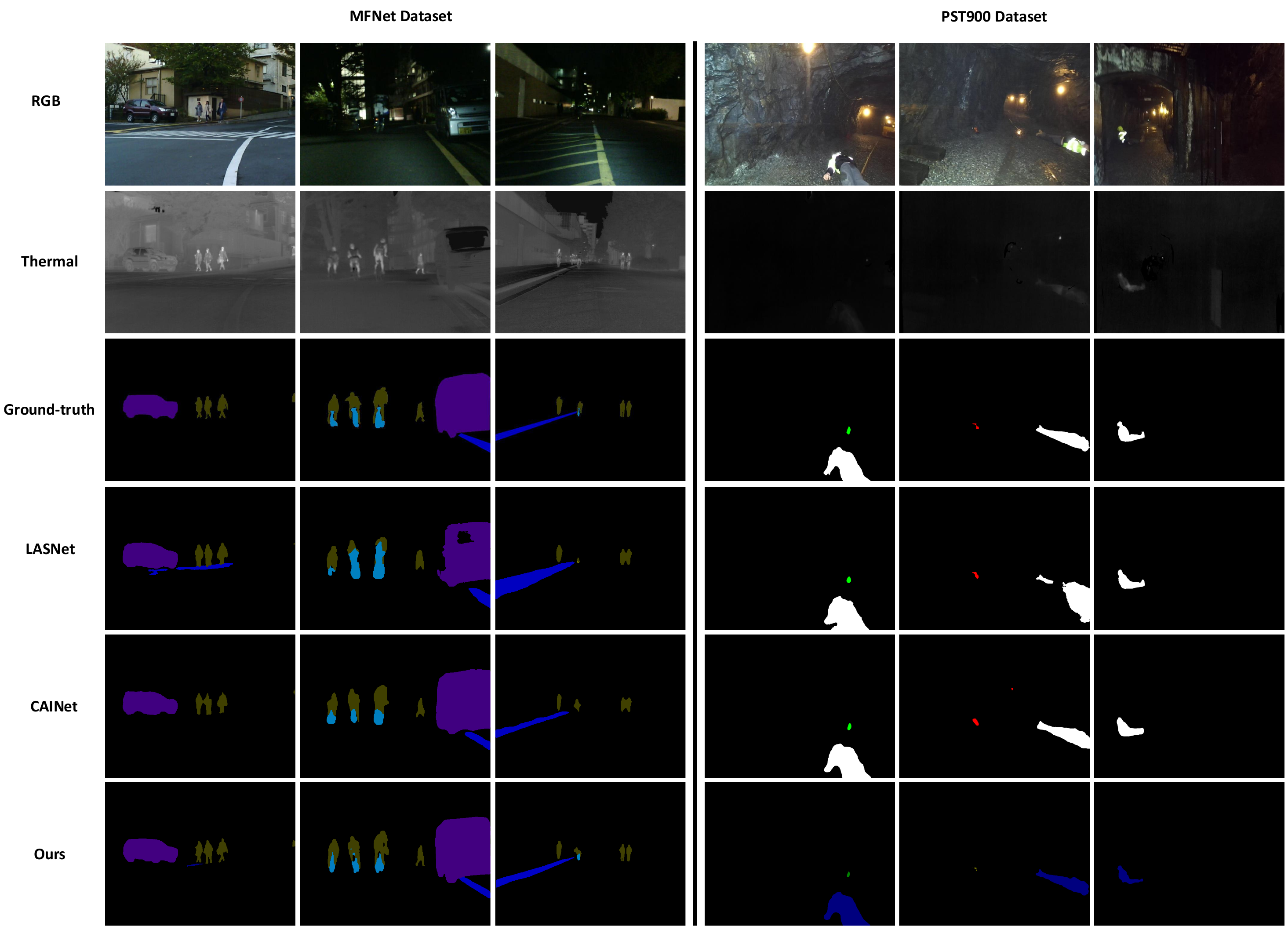}
\caption{Qualitative comparison on MFNet (Left) and PSD900 (Right) Datasets.}
\label{fig:thermal_vis}
\end{figure*}

\Cref{fig:thermal_vis} shows qualitative comparisons that further corroborate the numerical results. On both MFNet and PST900, our model produces clean, semantically accurate masks with sharp boundaries and high consistency with ground truth. It exhibits clear advantages in detecting small and low-visibility targets, especially under nighttime and occluded conditions, where other models, such as LASNet and CAINet, tend to fail or produce fragmented masks.

\vspace{0.8em}
\noindent\textbf{RGB-D Experiments.} Quantitative results on NYU Depth V2~\cite{Silberman2012} and SUN RGB-D~\cite{SunRGBD} datasets are shown in~\Cref{tab:depth}. Our approach, using the VMamba-S backbone, achieves an mIoU of \textbf{56.8} on NYU Depth V2, slightly behind CMNeXt (MiT-B4), and achieves the highest mIoU of \textbf{52.1} on SUN RGB-D among all tested models. This indicates strong cross-dataset performance, especially in indoor scene understanding, where spatial layout and fine structural details are critical.

\begin{figure*}[!t]
\centering
\includegraphics[width=0.99\linewidth]{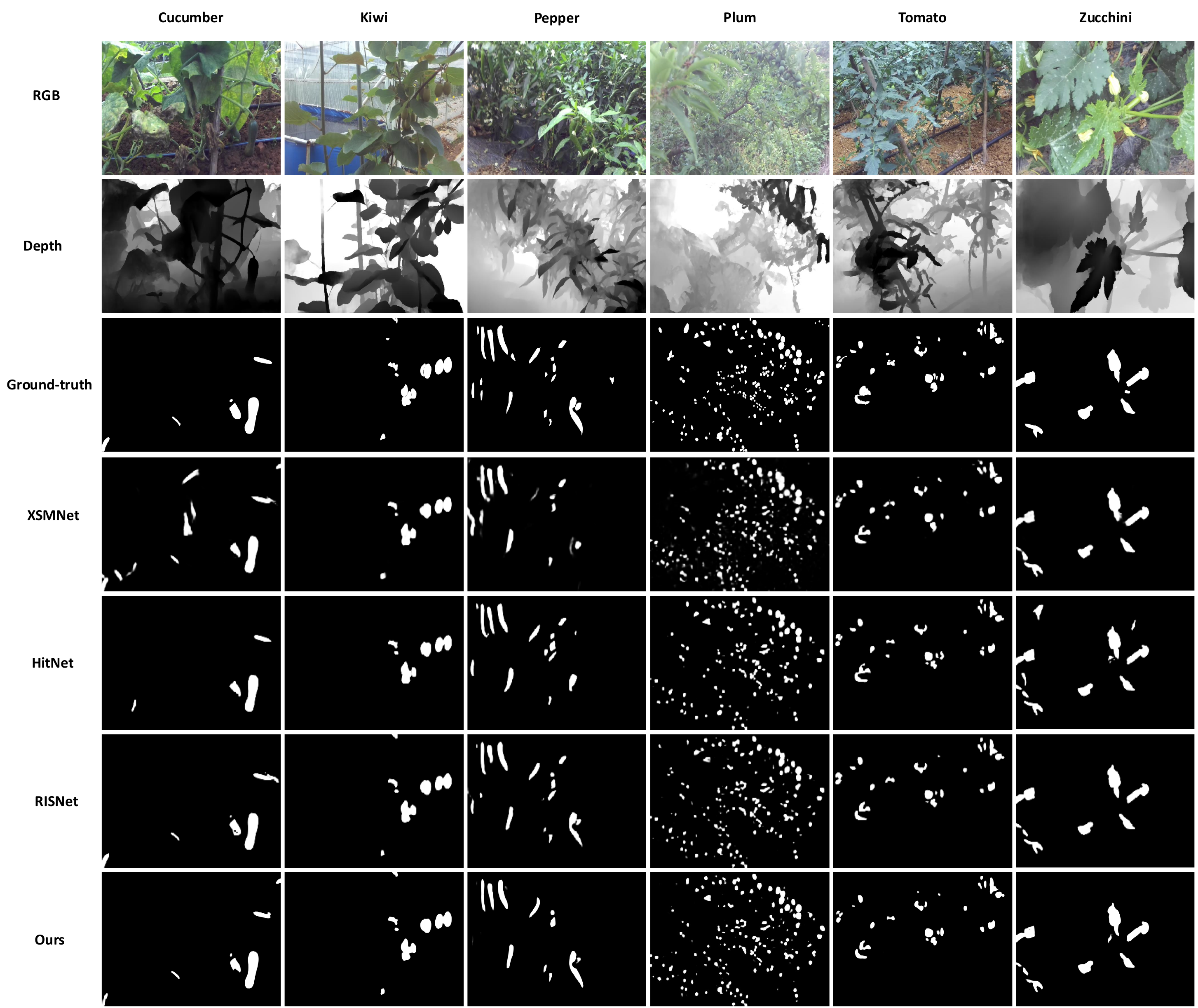}
\caption{Qualitative comparison on ACOD-12K Dataset.}
\label{fig:depth_agri}
\vspace{-0.4cm}
\end{figure*}

On the ACOD-12K dataset~\cite{Wang_2024_CVPR} (\Cref{tab:depth_agri}), which focuses on agricultural object detection under occlusion and clutter, our model significantly outperforms existing methods across all metrics: $S_\alpha = 0.865$, $F_{\beta}^w = 0.807$, and $E_\phi = 0.965$. Compared to RISNet, XSMNet, and HitNet, our method shows better spatial consistency and object delineation, as shown in~\Cref{fig:depth_agri}. Our model exhibits fewer false positives and better adherence to object boundaries across diverse crops like cucumber, pepper, and tomato.

\vspace{0.8em}
\noindent\textbf{Discussion.} Our method exhibits minor variation across metrics and datasets. These differences primarily reflect \ding{182} how individual metrics weight specific qualities (e.g., boundary precision vs. region consistency), \ding{183} dataset characteristics such as object scale and modality noise, and \ding{184} baselines that are more heavily tuned to particular datasets. Overall, our model remains consistently competitive across benchmarks, and the observed fluctuations are not statistically significant in most cases.

\begin{table*}[!htb]
\small
\begin{minipage}[t]{.56\linewidth}
\centering
\caption{\small Quantitative Comparisons on the NYU Depth V2 and SUN RGB-D Datasets (RGB-Depth).}
\label{tab:depth}
\centering
\setlength{\tabcolsep}{3pt}
\renewcommand{\arraystretch}{1.5}
\resizebox{\linewidth}{!}{%
\begin{tabular}{@{}c|c|cc|cc@{}}
\toprule
\multirow{2}{*}{\textbf{Method}} & \multirow{2}{*}{\textbf{Backbone}} & \multicolumn{2}{c|}{\textbf{NYU Depth V2}} & \multicolumn{2}{c}{\textbf{SUN RGB-D}} \\ \cmidrule(l){3-6} 
                                 &                                    & \textbf{In. Size}     & \textbf{mIoU}$\uparrow$    & \textbf{In. Size}   & \textbf{mIoU}$\uparrow$  \\ \midrule \midrule
CEN~\cite{wang2020deep}                              & ResNet-152                         & $480\times640$         & 52.5             & $530\times730$       & 51.1           \\
TokenFusion~\cite{wang2022multimodal}                      & MiT-B3                             & $480\times640$          & 54.2             & $530\times730$       & 51.0           \\
MultiMAE~\cite{bachmann2022multimae}                         & ViT-B                              & $640\times640$          & 56.0             & $640\times640$       & 51.1           \\
PDCNet~\cite{yang2023pixel}                           & ResNet-101                         & $480\times480$         & 53.5             & $480\times480$        & 49.6           \\
CMNeXt~\cite{zhang2023delivering}                           & MiT-B4                             & $480\times640$          & 56.9             & $530\times730$       & 51.9           \\
CAINet~\cite{lv2023cainet}                           & MobileNet-V2                       & $480\times640$          & 52.6             & -                     & -              \\
CMX~\cite{zhang2023cmx}                           & MiT-B4                       & $480\times640$          & 56.0            & $480\times640$                     & 52.1              \\
Sigma~\cite{wan2025sigma}                           & 	VMamba-S                       & $480\times640$          & 57.0             & $480\times640$                     & 52.4              \\
Ours                             & VMamba-S                           & $480\times640$         & 56.8              & $480\times640$        & 52.1            \\ \bottomrule
\end{tabular}
}
\end{minipage}%
\hfill%
\begin{minipage}[t]{.42\linewidth}
\centering
\caption{\small Quantitative Comparisons on the ACOD-12K Dataset (RGB-Depth).}
\label{tab:depth_agri}
\centering
\setlength{\tabcolsep}{3pt}
\renewcommand{\arraystretch}{1.5}
\resizebox{\linewidth}{!}{%
\begin{tabular}{@{}c|c|ccc@{}}
\toprule
\textbf{Method} & \textbf{Backbone} & $\bm{S_{\alpha} \uparrow}$ & $\bm{F_{\beta}^w \uparrow}$ & $\bm{E_{\phi} \uparrow}$ \\ 
\midrule \midrule
DaCOD~\cite{wang2023depth}          & Swin-L               & 0.80                          & 0.71                           & 0.91                        \\
PopNet~\cite{wu2023source}         & ResNet-50
               & 0.84                          & 0.78                           & 0.96                        \\
HitNet~\cite{hu2023high}         & PVT-v2-B2               & 0.85                          & 0.79                           & 0.96                        \\
FSPNet~\cite{huang2023feature}         & ViT-B               & 0.72                          & 0.53                           & 0.82                        \\
HIDANet~\cite{wu2023hidanet}        & R2Net-v1b-101               & 0.82                          & 0.73                           & 0.95                        \\
XMSNet~\cite{wu2023object}         & PVT-v2-B5               & 0.84                          & 0.75                           & 0.96                        \\
RISNet~\cite{Wang_2024_CVPR}         & PVT-v2-B2               & 0.87                          & 0.80                           & 0.97                        \\
FMamba~\cite{xie2024fusionmamba}         & 	VMamba-S               & 0.84                          & 0.55                           & 0.96                        \\
Ours           & VMamba-S          & 0.87                            & 0.81                             & 0.97                          \\ 
\bottomrule
\end{tabular}
}
\end{minipage} 
\end{table*}

\section{Conclusion and Future Works}
\label{sec:conclusion}

In conclusion, we introduce HiddenObject, a multimodal detection framework that integrates a Mamba-based fusion mechanism to address challenging scenarios involving occlusion and concealment—conditions under which traditional RGB imaging methods often fail. Our approach is evaluated across multiple datasets focused on thermal and depth modalities, with an emphasis on environments where object visibility is severely limited. As shown in the experiment, the proposed architecture demonstrates strong performance and generalizability, validating its effectiveness in detecting hidden objects under diverse conditions.

\vspace{0.8em}
\noindent\textbf{Limitations and Future Work.} We found that integrating Mamba with multimodal fusion for detecting hidden or partially concealed objects presents several challenges. These challenges include managing diverse data types, ensuring proper alignment between different modalities, and maintaining computational efficiency. Future research could investigate adapting Coupled Mamba for vision tasks, utilizing its state chain coupling to improve fusion in cooperative perception systems.

\vspace{0.8em}
\noindent\textbf{Broader Impact.} 
The HiddenObject framework has substantial potential to revolutionize multiple sectors by enhancing the detection of concealed or obscured objects across various real-world scenarios. In the domains of security and surveillance, enhanced multimodal detection capabilities afford more dependable threat identification, thereby supporting critical decision-making processes. Within industrial automation, advanced hidden object detection can significantly augment operational efficiency, minimize human errors, and enhance safety standards. Furthermore, the proposed framework markedly improves search-and-rescue operations by enabling more effective localization and recovery of objects or individuals in complex, visually challenging environments. In agriculture, our modality-agnostic fusion framework holds significant implications for precision farming and automated harvesting. By efficiently detecting fruits, vegetables, and other agricultural products, even when partially or fully concealed by foliage or densely clustered, the HiddenObject system can significantly enhance yield estimation accuracy, minimize waste, and aid in automating harvesting processes. Furthermore, improved detection capabilities facilitate real-time crop health monitoring and precision weeding or pruning operations, ultimately boosting agricultural productivity and sustainability. Thus, our approach not only addresses critical agricultural challenges but also encourages wider adoption of intelligent automation technologies in modern farming practices.

\vspace{0.8em}
\noindent\textbf{Licenses.} We conducted our experiments on different datasets. Therefore, we will follow their Terms of Use or Licenses. The copyright remains with the original owners of the datasets. In addition, our paper is under the \href{https://creativecommons.org/licenses/by/4.0/legalcode.en}{CC-BY-4.0} license, and it shall be used only for non-commercial research and educational purposes.

\ifpeerreview \else
\section*{Acknowledgments}
This work was supported in part by the US Department of Agriculture (grant numbers 2024-67021-42528 and 2022-67022-37021).
\fi

\bibliographystyle{IEEEtran}
\bibliography{references}








\end{document}